\documentclass{article}

\usepackage{arxiv}

\usepackage[utf8]{inputenc} 
\usepackage[T1]{fontenc}    
\usepackage{hyperref}       
\usepackage{url}            
\usepackage{booktabs}       
\usepackage{amsfonts}       
\usepackage{nicefrac}       
\usepackage{microtype}      
\usepackage{amsmath}        
\usepackage{multirow}        
\usepackage{siunitx}        
\usepackage{amssymb}        
\usepackage{threeparttable} 
\usepackage{lipsum}		    
\usepackage{graphicx}
\usepackage[square,numbers,sort&compress]{natbib}
\usepackage{doi}

\title{Polymorphic Combinatorial Frameworks (PCF): Guiding the Design of Mathematically-Grounded, Adaptive AI Agents}

\author{ \href{https://orcid.org/0009-0000-1518-3565}{\includegraphics[scale=0.06]{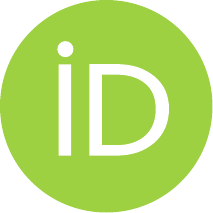}\hspace{1mm}David Pearl}\thanks{Corresponding author. Earlier version previously published on \href{https://www.researchsquare.com/article/rs-6397317/v1}{Research Square}.} \\
	Department of Mechanical Engineering\\
	Tufts University\\
	Medford, MA 02155 \\
	\texttt{david.pearl@tufts.edu} \\
	\And
	\href{https://orcid.org/0009-0006-4179-5526}{\includegraphics[scale=0.06]{orcid.pdf}\hspace{1mm}Matthew M. Murphy} \\
	Department of Mechanical Engineering\\
	Tufts University\\
	Medford, MA 02155 \\
	\texttt{mmurph31@tufts.edu} \\
    \And
	\href{https://orcid.org/0000-0002-5487-2715}{\includegraphics[scale=0.06]{orcid.pdf}\hspace{1mm}James Intrilgator} \\
	Department of Mechanical Engineering\\
	Tufts University\\
	Medford, MA 02155 \\
	\texttt{james.intriligator@tufts.edu} \\
}

\renewcommand{\headeright}{}
\renewcommand{\undertitle}{}
\renewcommand{\shorttitle}{Polymorphic Combinatorial Frameworks}

\hypersetup{
pdftitle={Polymorphic Combinatorial Frameworks (PCF): Guiding the Design of Mathematically-Grounded, Adaptive AI Agents},
pdfsubject={cs.AI, math.CO, stat.CO},
pdfauthor={David Pearl, Matthew Murphy, James Intriligator},
pdfkeywords={Combinatorial AI, Agent Design, Agents, In-Context Learning, Persona Multiplication, Polymorphism, Adaptability, Meta Prompting},
}

\begin{document}
\maketitle

\begin{abstract}
The Polymorphic Combinatorial Framework (PCF) leverages Large Language Models (LLMs) and mathematical frameworks to guide the meta-prompt enabled design of solution spaces and adaptive AI agents for complex, dynamic environments. Unlike static agent architectures, PCF enables real-time parameter reconfiguration through mathematically-grounded combinatorial spaces, allowing agents to adapt their core behavioral traits dynamically. Grounded in combinatorial logic, topos theory, and rough fuzzy set theory, PCF defines a multidimensional SPARK parameter space (Skills, Personalities, Approaches, Resources, Knowledge) to capture agent behaviors. This paper demonstrates how LLMs can parameterize complex spaces and estimate likely parameter values/variabilities. Using PCF, we parameterized mock café domains (five levels of complexity), estimated variables/variabilities, and conducted over 1.25 million Monte Carlo simulations. The results revealed trends in agent adaptability and performance across the five complexity tiers, with diminishing returns at higher complexity levels highlighting thresholds for scalable designs. PCF enables the generation of optimized agent configurations for specific scenarios while maintaining logical consistency. This framework supports scalable, dynamic, explainable, and ethical AI applications in domains like customer service, healthcare, robotics, and collaborative systems, paving the way for adaptable and cooperative next-generation polymorphic agents. See GitHub \href{https://github.com/cadavid1/PCF}{here} for experimental data and analysis tools.\end{abstract}

\keywords{Combinatorial AI \and Agent Design \and Agents \and In-Context Learning \and Persona Multiplication \and Polymorphism \and Adaptability \and Meta Prompting}

\section{Introduction}
Interest is growing in AI agents, whether computational, simulated, or physically embodied, that adapt to complex, evolving environments\cite{xi_rise_2025}. Despite advances, LLMs and AI frameworks often rely on static configurations, limiting dynamic adaptability \cite{pacheco_role_2003, zhou_symbolic_2024}. While specialized agents may address specific tasks, the broader challenge lies in designing systems that can autonomously evolve their configurations and capabilities to meet diverse and changing demands. For example, consider an AI assistant handling a medical emergency call. In seconds, it must shift from empathetic listener gathering symptoms, to analytical diagnostician, to clear instructor guiding first aid. It does this shifting not through pre-scripted responses, but by dynamically reconfiguring its core behavioral parameters. This is the promise of polymorphic agents, and the Polymorphic Combinatorial Framework (PCF) makes it possible today.

Here we introduce the Polymorphic Combinatorial Framework (PCF), an agentic architecture that builds upon Persona Multiplication (PM)\cite{pearl_persona_2023}, itself presenting a method for persona creation/examination. The PM method helps designers avoid designed injustice by examining multi-dimensional combinatorial spaces. The PM method addresses an important concern: because "\textit{personas determine destiny}," any project that launches with inaccurate (or too limited) persona(e/s) can lead to artifacts that don't meet all user group needs - thus leading to “inadvertent designed injustice”. Complication Matrices in PM explore outcomes across dimensions (e.g., attributes, roles, scenarios, tasks). Part of the art in design is in selecting relevant, appropriate, or desired dimensions to explore. PM leveraged this approach as part of its broader method aimed at helping designers navigate myriad possibility-spaces. This approach helps designers, or design teams, identify biases or (dis)advantaged groups. By considering persona-as-destiny, designers can ensure they are designing for the right sets of personas - and thereby design systems that avoid designed injustice. Here we use a similar concept to explore a different design-space. We use the same logic in PCF to systematically explore \textit{agents-space}. We leverage LLMs to identify, parameterize, and explore relevant possibility-spaces for\textit{ agents}. Here we recognize: in the world of LLMs(and especially in agentic approaches) ‘parameterization is destiny’.

The first step in the PCF framework (see Figure 4, Stage 1) is to consider a space (or environment, task, role, etc) and then to use LLMs to identify parameterizations (and estimate values and variabilities) that best capture the range of agents operating in that space. Here, we used Claude to parameterize agents operating in simulated dining establishments and identify viability thresholds under suboptimal conditions. While prompts and outputs from LLMs like Claude are inherently stochastic, they allow for dynamic adaptability, reflecting real-world variability in complex systems. See supplementary materials for representative Python scripts and experimental data. For example, an LLM might be asked: "Describe a dining establishment operating at minimal viability, considering factors such as deteriorating infrastructure, disadvantaged geographical placement, and clientele diversity. After describing the establishment, identify the primary dimensions (or variables, attributes, etc.) that can best quantify the agents operating in this establishment and also which can be used to distinguish this establishment's agents from others. Finally, offer estimates of parameter values and variabilities that best capture agents in these types of establishments."  It is worth noting that more accurate parameterizations (and values/variabilities) can be achieved through various techniques, such as richer prompting, RAGs, etc. 

We iteratively analyzed the LLM outputs to identify consistent parameterization patterns across multiple runs, with stochasticity handled through parameter averaging and qualitative validation. While exact reproducibility is constrained by the nature of LLMs, the derived parameters can be shared for replication efforts. This approach balances the strengths of dynamic LLM-driven insights with the challenges of repeatability, acknowledging that the adaptability of LLMs, while a limitation for deterministic replication, mirrors the variability of real-world environments. An additional advantage of this approach is that it can bake explainability right into the process from the very beginning. The parameters can be generated by LLMs to align with explainable/understandable human concepts and terms.

The second primary step in PCF (Figure 4, Stages 2 and 3) is to use the parameterizations from Step 1 to create agents to operate in the parameterized environment. The agents are crafted by giving them sets of Skills, Personalities, Approaches, Resources, and Knowledge (SPARKs). By doing this, PCF equips agents to adapt dynamically via in-context learning and reparameterization of these SPARK configurations. The SPARK parameter space defines multidimensional attributes governing agent behavior: Skills reflect task-specific abilities (e.g., cooking or customer service), Personalities shape interpersonal dynamics (e.g., accommodating vs. assertive), Approaches define methods (e.g., teamwork vs. independent execution), Resources include tools and/or available abstract or concrete assets (money, LLM routing agent, etc), and Knowledge represents domain expertise or databases. 

For instance, a customer service agent might combine Skills(technical support=8, empathy=7), Personality(patient=9, assertive=4), Approach(collaborative=8), Resources(knowledge base=access, escalation protocol=enabled), and Knowledge(product specs=expert, company policies=current).

It is important to note the role of ‘Goals’ within the PCF methodology. While not included as a combinatorial parameter within SPARK, Goals function as the objective criteria used to evaluate the fitness of any given agent configuration. They are the external, often dynamic, success or optimization metrics (e.g., maximize customer satisfaction, minimize response time, reduce carbon footprint) that guide the selection of an optimal SPARK configuration from the vast possibility space. 

In our example of a simulated café environment, agents optimize SPARK parameters for roles like chef, waiter, or sommelier. A chef might emphasize Skills such as cooking speed and Resources like recipe access, while a sommelier prioritizes Knowledge of wine pairings and Personality traits like charisma. By adjusting SPARK configurations, agents shift seamlessly between roles to align with evolving contextual demands, ensuring adaptability beyond static designs.

PCF agents detect reconfiguration needs through multiple mechanisms: conversation classifiers identifying phase transitions ('symptom gathering' to 'diagnosis'), performance monitors detecting goal misalignment (satisfaction scores dropping), explicit user signals ('I need technical details'), or contextual triggers (detecting urgency keywords like 'emergency'). The specific triggering logic depends on implementation, a medical AI might use symptom severity classifiers, while customer service uses sentiment analysis. Critically, the LLM itself can serve as the meta-controller, evaluating conversation context and determining when SPARK adjustments would improve outcomes. This keeps reconfiguration decisions interpretable and auditable.

The third primary step in PCF is to ensure that any spawned agent (i.e. SPARK configuration) does not contain any contradictory or impossible combinations of attributes (see Figure 4, Stage 3). Grounded in topos theory and rough fuzzy sets, PCF ensures consistent, non-contradictory configurations (e.g., 'helpful' vs. 'obstructive'). To validate PCF, we used an LLM\cite{nair_further_2007} to envision, create, and parameterize, simulated café environments of different complexity levels. We then performed Monte Carlo simulations(see Figure 4, Stage 4) where the dimensions simulated – as well as their mean values and variabilities - were created using the LLM\cite{nair_further_2007}. We generated 1.25 million data points allowing us to analyze agent adaptability across varying café complexity levels. These simulations demonstrated PCF’s ability to manage variability while maintaining robust performance. This paper explores PCF’s synergies and transformative potential, from Explainable AI (XAI) to Reinforcement Learning (RL) and human-AI collaboration. By combining rigorous mathematical principles with practical adaptability, PCF provides a foundation for scalable, flexible, and ethically-aligned AI systems capable of addressing real-world challenges. 

\section{Results}

\subsection{Agent Prompt Construction}

For readers not ready to dive into the math just yet, here’s the gist: We use LLMs to generate plausible agent configurations by combining elements from the SPARK space: Skills, Personalities, Approaches, Resources, and Knowledge. Then, using formal methods, we systematically prune out combinations that are logically inconsistent, contradictory, or unfit for a given context. What remains is a structured space of viable agent designs, each coherent, adaptive, and ready for simulation or deployment. This section walks through the math that makes this possible. 

During inference, LLM outputs vary stochastically – producing qualitatively and quantitatively different results across parameter combinations\cite{bender_dangers_2021, selig_improving_2012, ramalho_simulation_2013}. This variability often approximates a normal distribution, arising from the renormalization functions intrinsic to LLM architectures\cite{bar_massada_incorporating_2008}. Within PCF, hyper-parametric modeling leverages this variability to predict (and/or help set) output distributions in simulations or to iteratively improve agents. Using set theory, we define the possibility space of agent parameters and configurations, systematically exploring all mathematically valid combinations of Skills, Personalities, Approaches, Resources, and Knowledge (SPARK) to identify contextually relevant and operationally effective configurations within a particular task/context space.

We can define the set of all possible configurations (see Figure 4, Stage 2)  $S_{Possibility}$ as: 

$S_{Possibility} = {Skills} \times {Personalities} \times {Approaches} \times {Resources} \times {Knowledge}$

In this case, $S_{Possibility}$ encompasses all possible parameter configurations, defining the possibility space of solutions based on the specific use case, context, and parameterization. When a particular parameter must be fixed (e.g., specifying a particular Approach), it becomes essential to examine the subset of configurations for the remaining parameters. 
This subset, $S_{Plausibility}$, is defined as $S_{Possibility}$ as constrained by (or intersected with) the context(s) and use case(s) (the domain constraints) configured:
$S_{Plausibility} = S_{Possibility} \cap \mbox{Domain Constraints}$. 
For example, if we decide to set the Approach $a$, then we can define the subset of all plausibilities with this particular value of $a$ specified, $S_a$, as:

$S_a = \{ (skills,personalities,approaches,resources,knowledge) \mid s \in Skills, p \in Personalities, r \in Resources, k \in Knowledge \}$

By analyzing the combinatorial structure of the plausibility space for $S_{Plausibility}$, we mathematically ensure logical consistency. This ensures that non-viable parameter combinations, such as mutually exclusive traits, are excluded, resulting in agent configurations that are both rationally coherent and contextually relevant.

In the PCF, sets anchored by specific use cases or contexts provide prompt engineers with a diverse and adaptable plausibility space for SPARK parameterization. However, not all parameter combinations are valid; for example, assigning an agent both {personality=helpful} and {approach=obstructive} represents poor reasoning. Such contradictions render the agent unable to fulfill its role effectively and may bias outputs toward pre-trained data\cite{pryzant_automatically_2020}. To address this, we employ topos theory\cite{baez_topos_2021} to systematically analyze these sets and ensure logical consistency in agent configurations.

In category theory, the relationships between objects (SPARK nodes) are formalized as morphisms, which represent structured transformations or mappings\cite{david_i_spivak_category_2014,tull_towards_2024}. Within PCF, functors, which are mappings between categories, serve as "rules of engagement," guiding how objects and morphisms interact across categories. By carefully structuring these functors, we can translate morphisms between categories, restrict incompatible choices within each category, and exclude impractical configurations. This ensures that agent parameters are both logically consistent and operationally effective within their intended use cases.
By integrating set theory and category theory, PCF provides a robust mathematical foundation for managing agent configurations. This integration allows for the quantification of the plausibility space, the systematic elimination of contradictory configurations, and the facilitation of adaptive prompt engineering (e.g. parameterizing more than one personality trait is possible, but it is illogical to parameterize both "generous" and "stingy" within an agent’s design). Together, these methods underpin the development of agents (computational, physical, etc.) that are powerful, flexible, coherent, and reliable, ensuring they meet the demands of their environments.

\subsection{Experimental Data Analysis}

To evaluate the effectiveness of PCF, we developed café simulations of varying levels of complexity. Simulations modelled the performance of servers serving diners. Across conditions, servers were either adaptable (using PCF approaches) or non-adaptable. The simulation included five tiers of cafés, with ratings from one-star to five-star, with increasing complexity in environmental factors, resource availability, and service quality. Initially, we employed Claude\cite{anthropic_introducing_nodate} to help parameterize the entire simulation space – identifying relevant variables, dimensions, etc. We then employed Claude to inform the simulation's parameter values – including means, medians, and measures of variance. Subsequently, we used it to provide context-aware estimates of key variables such as customer expectations, service times, and satisfaction scores. Claude-derived values served as inputs for the simulation, enabling a realistic representation of agent-customer interactions across varying conditions.

In this study, we chose to utilize Claude due to its strong multi-turn conversational coherence and accessibility at the time of experimentation. Future research can explore the implementation and efficacy of PCF across a wider set of foundation and open-source language models such as GPT, LLaMA, or smaller open-source alternatives. Future work should evaluate the performance of PCF with a range of models, including smaller and larger architectures, to better understand the impact of model size on adaptability within the SPARK parameter space. Additionally, comparative analyses incorporating human feedback and emerging techniques like Test-Time Compute (TTC) and Self-Play could further refine PCF's robustness and scalability. While these investigations remain beyond the scope of this study, they represent crucial directions for enhancing the framework’s applicability across diverse contexts.

The simulation generated 1.25 million data points, capturing customer experiences across five tiers. We conducted an Ordinary Least Squares (OLS) regression to analyze the relationship between customer satisfaction ($satisfaction\_score$) and total time per meal ($total\_time\_per\_meal$). The results of this analysis are summarized in Table 1 below.

\begin{table}[htbp]
\centering
\resizebox{\textwidth}{!}{%
\begin{threeparttable}
\small
\caption{OLS Regression Results}
\label{tab:ols_results}
\begin{tabular}{lcccccc}
\toprule
\textbf{Variable} & 
\textbf{Coefficient} & 
\textbf{Std. Error} & 
\textbf{t-statistic} & 
\textbf{p-value} & 
\textbf{95\% CI (Lower)} & 
\textbf{95\% CI (Upper)} \\
\midrule
Intercept          & 2.9239 & 0.022 & 134.440 & 0.000 & 2.881 & 2.967 \\
satisfaction\_score & 2.3370 & 0.004 & 589.442 & 0.000 & 2.329 & 2.345 \\
\bottomrule
\end{tabular}
\begin{tablenotes}\footnotesize
\item \textbf{Dependent Variable}: total\_time\_per\_meal
\item \textbf{R-squared}: 0.217, \textbf{Adjusted R-squared}: 0.217
\item \textbf{F-statistic}: 3.474e+05, \textbf{Prob (F-statistic)}: 0.00
\item \textbf{Log-Likelihood}: -3.9283e+06, \textbf{AIC}: 7.857e+06, \textbf{BIC}: 7.857e+06
\item \textbf{No. Observations}: 1250000, \textbf{Model DF}: 1, \textbf{Residual DF}: 1249998
\item \textbf{Omnibus}: 23274.940, \textbf{Prob(Omnibus)}: 0.000
\item \textbf{Jarque-Bera}: 24406.050, \textbf{Prob(JB)}: 0.00
\item \textbf{Skew}: 0.337, \textbf{Kurtosis}: 2.880, \textbf{Durbin-Watson}: 1.581
\end{tablenotes}
\end{threeparttable}%
} 
\end{table}

The regression analysis revealed a statistically significant positive relationship between $total\_time\_per\_meal$ and $satisfaction\_score$. Figures 1 and 2 below present graphical representations of these findings, illustrating the observed relationships and the distribution of meal durations across satisfaction scores. Each one-unit increase in satisfaction score was associated with an increase of approximately 2.34 minutes in meal duration. But, we observed that this was modulated by aspects of the PCF parameterization - for example, observe the large tail in the 1-star relationship. The high t-values and low p-values for the coefficients confirm the robustness of this result. The model’s r-squared value of 0.217 indicates that 21.7\% of meal duration variability can be attributed to customer satisfaction, suggesting that other factors not captured in this model also contribute to meal duration. Additionally, the upper and lower confidence intervals at 95\%, combined with standard errors, showed that the simulated data fell within expected ranges and remained statistically significant (p<0.05).

\begin{figure}
    \centering
    \includegraphics[width=1\linewidth]{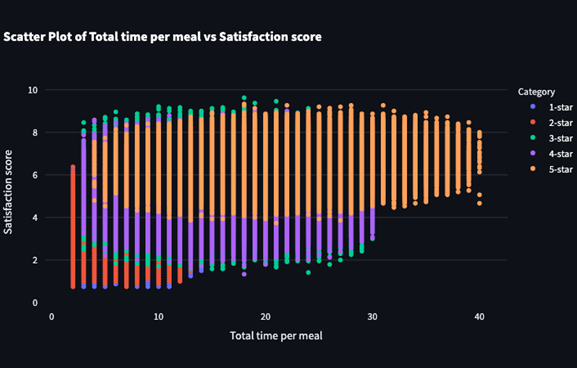}
    \caption{Scatter Plot of Total Time per Meal vs. Satisfaction Score}
    \label{fig1}
\end{figure}

\begin{figure}
    \centering
    \includegraphics[width=0.5\linewidth]{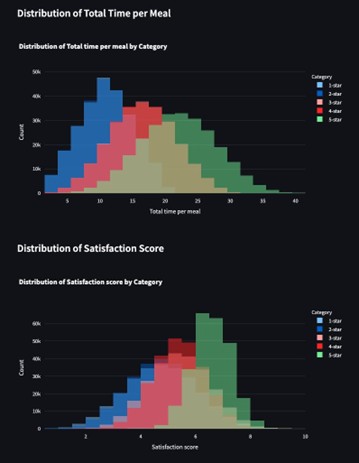}
    \caption{Distribution of Total Time per Meal Across Different Satisfaction Scores}
    \label{fig2}
\end{figure}

As shown in Figure 2,  the data exhibit variance patterns consistent with the polymorphic composition of the five simulated café tiers. Each tier's simulation parameters were associated with unique distributions, and the mean of these distributions predicted time and satisfaction according to normal distribution properties as expected. The curvature and visible plateau suggest that other factors, not captured within the current model, play a significant role in influencing customer behavior - this will be explored further below.

The 1-star café simulation, with its minimal resources and deliberately impeded parameters, functions as a de facto baseline against which the adaptability of more complex agent configurations can be measured. We observed a strong overlap between one star and two to three star performance, with a partial overlap between one and four stars. Claude identified instances of dining establishments operating at minimum viability thresholds - specifically, establishments characterized by suboptimal environmental conditions, including deteriorating infrastructure, disadvantaged geographical placement, and a clientele displaying marked interpersonal friction. Notably, while the parameters of the single-star simulation framework deliberately impeded customer satisfaction metrics, the PCF adaptive mechanism nevertheless generated emergent patterns of satisfaction. This unexpected emergence of favorable outcomes despite adverse initial conditions demonstrates the robust nature of complex adaptive systems in applied AI. The PCF mechanism appears to facilitate self-organizing behaviors that transcend baseline operational parameters, suggesting implications for our understanding of customer satisfaction as an emergent property rather than a directly engineered outcome.

\subsection{Mathematical Formulation of Agents}

While the previous sections established the simulation architecture and parameterization logic of SPARK configurations, this next section pivots toward formal mathematical grounding. We develop a sheaf-theoretic framework, rooted in category and topos theory, to rigorously define the space of logically coherent agents.

Before engaging with the mathematical formalism in this section, it may help to clarify the core function it serves. Once SPARK configurations are generated, we need a principled way to ensure that no agent ends up with internally inconsistent traits, for example, one that combines a “helpful” personality with an “obstructive” approach.  This section introduces the formal machinery, grounded in category and topos theory, that allows us to define and enforce logical coherence across the multidimensional SPARK space. These methods ensure that only valid, interpretable agent configurations proceed to the simulation stage.

PCF can be applied in two primary configurations: Single-Agent Configuration and Multi-Agent Configuration, each governed by a distinct mathematical framework.

\subsubsection{Single-Agent Configuration}
The total number of possible configurations for a single agent is calculated as the sum of binomial coefficients, representing all possible combinations of ($Skills$), ($Personalities$), ($Approaches$), ($Resources$), and ($Knowledge$). Mathematically, this is expressed as:
$$C_1= \left( \sum_{k}^{n} \binom{Skills \times Personalities \times Approaches \times Resources \times Knowledge}{k} \right)$$
Where:
$n$ = number of total parameters
$k$ = number of parameters selected (from 1 up to $n$)
Explanation:
Binomial Coefficient $(\binom{n}{k})$: Represents the number of ways to choose $k$ parameters from $n$ available parameters.
Summation over $k$: Accounts for all possible numbers of parameters selected, from 1 up to $n$.
Example Calculation:
If we have ($Skills = 4$), ($Personalities = 3$), ($Approaches = 3$), ($Resources = 3$), and ($Knowledge = 2$), then:
$$C_1= \left( \sum_{k=1}^{n} \binom{4 \times 3 \times 3 \times 3 \times 2}{1} \right) =216$$

For this agent configuration-space, our example SPARK parameters can yield a single agent with a possibility space of up to 216 different combinations of parameters.

\subsubsection{Multi-Agent Configuration \texorpdfstring{$C(A)$}{C(A)}}

The total number of possible configurations for a set of $n$ agents is: \\
$C(A) = (S_\text{Possibility})^n$
This exponential growth highlights the vast potential for creating diverse and specialized agents within PCF. Each agent adds a new dimension to the total potential agent parameter space, multiplying the possibilities by its SPARK parameters.
The set of all possible configurations $S$ is given by the Cartesian product:
$$S = \bigcup_{k=1}^{n} \left( \binom{Skills \times Personalities \times Approaches \times Resources \times Knowledge}{k} \right)$$
Where $$\binom{Skills \times Personalities \times Approaches \times Resources \times  Knowledge}{k}$$ denotes all subsets of SPARK parameters containing $k$ elements.

\subsubsection{Eliminating Redundancies and Conflicts}
Since not all combinations in $S$ are useful or logically consistent, we define a novel subset, $S_{\text{valid}}$, that excludes contradictory configurations per domain-specific constraints $C$. In this context, a contradiction arises when two or more SPARK components jointly imply semantically or behaviorally incompatible traits, for example, an agent marked as both “maximally cautious” and “impulsively reactive.”  The formalism here is designed to detect and exclude such incoherent combinations.

$$S_{\text{valid}} = \{ s \in S \mid s \text{ satisfies all constraints in } C \}$$

\subsection{Category Theory in PCF}

This section presents the architecture and mathematics behind our simulation approach. For those primarily interested in the conceptual outcomes rather than the technical details, the core idea is this: once logically coherent agent configurations are defined, we subject them to large-scale stochastic simulation to examine how they perform across different levels of environmental complexity. This allows us to identify patterns of adaptability, resilience, and diminishing returns.

We develop a rigorous categorical framework for modeling multi-parameter agent configurations, based on the SPARK dimensions: Skills, Personalities, Approaches, Resources, and Knowledge. Each parameter is treated as a site, and the integration of these sites is realized through fibered products over a common base of instructional contexts. A sheaf defined on the resulting fibered site encodes the agent's globally coherent behavior, with covering conditions representing local specializations and gluing axioms ensuring integrated solutions. This sheaf-theoretic formulation provides a precise foundation for modeling distributed, multi-dimensional cognition in synthetic agents\cite{david_i_spivak_category_2014,tull_towards_2024,hassan_hyper-dimensional_2022, woolsey_combinatorial_2020,richard_jensen_computational_2008}.

\subsubsection{Parameter Spaces as Grothendieck Sites}

This section formalizes how we treat parameter spaces in PCF using Grothendieck topologies. The key insight is that we need mathematical rules for determining which agent traits can be meaningfully combined in specific contexts (locally meaningful configurations), and how these local combinations aggregate into globally coherent agents. For example, 'helpful personality + collaborative approach' works well locally in customer service, but we need formal methods to ensure this local coherence extends to the agent's overall behavior across all situations.

Each SPARK parameter domain $X \in \{S, P, A, R, K\}$ is modeled as a small category $C_X$, equipped with a Grothendieck topology $J_X$. These data define a site $(C_X, J_X)$.
Objects in $C_X$ represent abstract entities relevant to the parameter domain: e.g., a skill in $C_S$, or a personality archetype in $C_P$.
Morphisms in $C_X$ encode contextual transformations or refinement relations: e.g., a morphism in $C_A$ may represent strategic adaptation.
The Grothendieck topology $J_X$ defines, for each object $U \in C_X$, a collection of covering families $\{U_i \to U\}$, representing localized refinements that together constitute $U$.
These topologies formalize the principle of local representability: complex or abstract elements within a parameter domain can be described via collections of simpler, locally defined components.

\subsubsection{Contextual Integration via Fibered Products}

Here we apply fibered product structures to integrate multiple context-sensitive SPARK configurations. Conceptually, this allows us to align and reconcile agents operating in overlapping but distinct environments, ensuring they share consistent parameters where needed while still adapting to their local constraints.

Let $\mathcal{I}$ denote the base category of user instructions or task contexts. Each parameter category $C_X$ is equipped with a functor $p_X: C_X \to \mathcal{I}$, associating each parameter object to its contextual anchoring.
The total configuration category $C$ is constructed as the fibered product:
$C = C_S \times_{\mathcal{I}} C_P \times_{\mathcal{I}} C_A \times_{\mathcal{I}} C_R \times_{\mathcal{I}} C_K.$

An object in $C$ is a tuple $(s, p, a, r, k)$ such that each component maps to the same base instruction $i \in \mathcal{I}$. Morphisms in $C$ consist of componentwise morphisms that commute over the same morphism in $\mathcal{I}$.
A Grothendieck topology $J$ is defined on $C$ by declaring that a family $\{X_i \to X\}_{i \in I}$, with $X = (s,p,a,r,k)$, constitutes a cover if and only if for each parameter projection $\pi_X: C \to C_X$, the induced family $\{\pi_X(X_i) \to \pi_X(X)\}_{i \in I}$ is a cover in $(C_X, J_X)$. This ensures parameter-wise locality is preserved in the total space.

With this fibered construction in place, we now turn to a higher-level integration mechanism: sheaves. Sheaf theory enables us to coherently combine locally valid agent behaviors into globally interpretable models, which is crucial for reconciling heterogeneous configurations across role, task, and context.

\subsubsection{Sheaves and the Gluing of Parameter Contributions}

This section builds on earlier formalisms by introducing sheaves as a way to manage and reconcile locally defined SPARK parameters across contexts. In essence, we use sheaf-theoretic gluing to ensure that even when parameters are defined in fragmented or overlapping ways, for example, across different roles, agents, or environments, they can be integrated into a consistent, globally meaningful configuration. Readers unfamiliar with sheaf theory may wish to skim the formal development and focus on the conceptual role it plays in linking partial views into coherent agent models.

A sheaf $\mathcal{F}: C^{\mathrm{op}} \to \mathbf{Sets}$ assigns to each agent configuration $X = (s,p,a,r,k) \in C$ a set $\mathcal{F}(X)$ representing well-formed agent behaviors (e.g., valid solutions or outputs) under that configuration. The restriction maps encode the specialization of solutions as configurations vary.

The sheaf condition requires that for any cover $\{X_i \to X\}_{i \in I} \in J$, the following diagram is an equalizer:

$\mathcal{F}(X) \longrightarrow \prod_{i} \mathcal{F}(X_i) \rightrightarrows \prod_{i,j} \mathcal{F}(X_i \times_X X_j).$

This condition guarantees the global coherence of behavior across local variations: if local sections $\{s_i \in \mathcal{F}(X_i)\}$ are compatible on overlaps, then they uniquely determine a global section in $\mathcal{F}(X)$.

\subsubsection{Functorial Interpretation and Parameter Interoperability}

Projection functors $\pi_X: C \to C_X$ induce natural transformations between sheaves, facilitating the translation of local sections across parameter domains. Such functorial structures encode the agent's capacity to adapt or reinterpret behaviors across changing configurations while maintaining semantic integrity.

For example, morphisms in $C_P$ can transform a solution from one personality profile into another, preserving its core structure while recontextualizing its presentation. This formalizes adaptive behavior within the agent's representational system.

\subsubsection{Synthesis of the SPARK Sheaf Model}

Let $C$ denote the fibered product category over $\mathcal{I}$ of the parameter categories $C_X$. Let $J$ denote the Grothendieck topology defined by componentwise covers. Then a sheaf $\mathcal{F}: C^{\mathrm{op}} \to \mathbf{Sets}$ satisfies:
\begin{itemize}
    \item $\mathcal{F}(X)$ is the space of agent outputs for configuration $X = (s,p,a,r,k)$;
    \item Restriction maps reflect parameter specialization;
    \item Covers $\{X_i \to X\}$ encode localized agent configurations;
    \item The equalizer condition ensures coherence of gluing.
\end{itemize}
Formally:
$$
\mathcal{F}(X) \cong \left\{\{s_i \in \mathcal{F}(X_i)\}_{i \in I} \;\middle|\; s_i|_{X_i \cap X_j} = s_j|_{X_i \cap X_j} \;\forall i,j\right\}.
$$
The sheaf gluing condition is expressed as an isomorphism ($\cong$) between:
\begin{itemize}
    \item The global section $\mathcal{F}(X)$, i.e., the agent’s coherent behavior under configuration $X = (s,p,a,r,k)$,
\end{itemize}
and
\begin{itemize}
    \item A compatible family of local behaviors ${s_i}$, each defined on a sub-configuration $X_i$, such that all overlaps $X_i \cap X_j$ agree, i.e., the behaviors $s_i$ and $s_j$ are consistent wherever they intersect.
\end{itemize}

This equation encodes the idea that local solutions across dimensions (skills, personas, etc.) can be glued into a global, coherent behavior, if and only if they agree on how they handle shared sub-tasks.

Overview of Key Components
\begin{itemize}
    \item $\mathcal{F}(X)$: The sheaf evaluated at $X$; think of this as the agent’s response or behavior when configured with parameters $s, p, a, r, k$.
    \item ${Xi \to X}{i \in I}$: A covering family, i.e., a collection of more specialized configurations that together 'cover' the full configuration $X$.
    \item $\mathcal{F}(X_i)$: The behaviors the agent can exhibit under configuration $X_i$, which is simpler or more localized than $X$.
    \item $\mathcal{F}(X_i \cap X_j)$: The behaviors over the intersection of two configurations $X_i$ and $X_j$. These must match for consistency.
    \item $\left\lbrace s_i\right\rbrace$: Local behaviors that we wish to glue.
    \item Condition $s_i|{X_i \cap X_j} = s_j|{X_i \cap X_j}$: Ensures pairwise compatibility on overlaps; if two local views agree wherever they overlap, they can be assembled into a unique whole. 
    
\end{itemize}

This framework ensures that agent responses integrate multidimensional parameter contributions in a coherent, context-sensitive, and mathematically principled manner. This sheaf-theoretic logic not only guarantees local-to-global consistency but also sets the stage for integrating additional mathematical paradigms that address fuzziness, polymorphism, and inferential noise in LLM-mediated behaviors.

\subsection{Integration of Mathematical Frameworks}

PCF (Polymorphic Combinatorial Framework) is best understood as a meta-prompting infrastructure, a system of parameter spaces  that defines how agent behaviors are expressed, organized, and adapted dynamically by LLMs. Rather than existing solely as abstract mathematical constructs, the SPARK parameter space (Skills, Personality, Approach, Resources, Knowledge) functions as a series of open variables within a generative system of agent configuration. These variables are interpretable and manipulable directly through prompts, enabling LLMs or meta-agents such as an "Agent Builder" to instantiate context-sensitive agents dynamically across problem spaces like administration, logistics, pedagogy, and compliance \cite{richard_jensen_computational_2008,fatima_rough_2024,halmos_naive_1974,julien_chaumond_llm_2024,mingard_deep_2025}.

Two primary mathematical frameworks undergird this parameter system. First, topos theory provides a categorical structure for organizing and composing agent designs within a coherent logic. In PCF, each parameter category (e.g., Skills or Approaches) corresponds to a site over which sheaves of agent configurations are defined. This ensures consistency under refinement: local parameter selections (e.g., for one department or individual) yield globally coherent agent behaviors under gluing conditions, governed by a Grothendieck topology \cite{baez_topos_2021, jia_category-theoretical_2024}. The framework supports polymorphism by enabling agent representations to transform via functorial mappings while preserving their structure, a critical property for systems in which one agent configuration must evolve into another based on context.

Second, rough fuzzy set theory models the uncertainty inherent in both heuristic selection and probabilistic inference by LLMs. Since these models operate as non-deterministic approximators, their outputs are not crisp classifications but contextually modulated estimations. Under this paradigm, LLMs act as rough fuzzy set classifiers, operating in upper and lower approximations of conceptual boundaries in SPARK parameter space \cite{richard_jensen_computational_2008,fatima_rough_2024, halmos_naive_1974, julien_chaumond_llm_2024,mingard_deep_2025}. When an agent is requested (e.g., "Create a training agent for senior staff"), the model does not select a static archetype, but instead interpolates from nearby agent representations based on context and feedback loops\cite{nezhurina_alice_2024}.

Together, these components enable a new class of adaptive, generative agents: parameterizable, composable, and semantically aligned with their tasks. Topos theory ensures that compositional logic remains intact across dynamic transformations, while rough fuzzy logic accounts for the ambiguity of natural language instructions and stochastic inference in LLM-based systems \cite{mingard_deep_2025,gu_survey_2025}. The result is a meta-prompting framework where LLMs can reason about, mutate, and orchestrate agent clusters across any domain, not by reprogramming agents, but by updating the parameter sheaves that define them.  

\subsection{Empirical Validation and Implications}

PCF significantly boosts LLM performance, as shown by both robust quantitative outcomes and richer qualitative responses. In our Monte Carlo Café simulations, higher PCF parametric complexity correlated with improved agent outcomes. Alongside linear OLS regression, a non-linear spline model captured nuanced trends in how meal duration influences satisfaction across café tiers, patterns a simple linear analysis might miss.
For upscale cafés, longer service can significantly raise satisfaction until hitting a threshold where further extension yields modest gains. By contrast, small increments of service in lower-tier cafés can quickly enhance the experience, reflecting how context dictates outcomes. These results affirm the PCF core idea that “more” is not automatically “better.” Both Table 1 (the earlier OLS model) and nonlinear (Figure 3 and Table 2) findings reveal a “sweet spot” shaped by resource constraints and trade-offs, where tuning SPARK parameters effectively heightens outcomes without undue costs.

\begin{figure}
    \centering
    \includegraphics[width=1\linewidth]{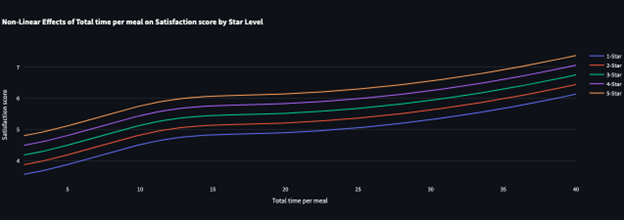}
    \caption{Non-Linear Effects of Total time per meal on Satisfaction Score by Star Level}
    \label{fig3}
\end{figure}

\begin{table}[!ht]
\centering
\begin{threeparttable}
\caption{Non-Linear Model Summary}
\label{tab:nonlinear_spline_regression}

\begin{tabular}{
    l
    S[table-format=1.4]  
    S[table-format=1.3]  
    S[table-format=3.3]  
    S[table-format=1.3]  
    l                    
}
\toprule
\textbf{Parameter} & \textbf{Coef} & \textbf{Std Err} & \textbf{t} & \textbf{p-value} & \textbf{95\% CI} \\
\midrule
Intercept & 3.1489 & 0.024 & 130.702 & 0.000 & [3.102, 3.196] \\
Star\_Level & 0.3122 & 0.002 & 146.043 & 0.000 & [0.308, 0.316] \\
bs(total\_time\_per\_meal, degree=3, df=5)[0] & 0.4011 & 0.038 & 10.609 & 0.000 & [0.327, 0.475] \\
bs(total\_time\_per\_meal, degree=3, df=5)[1] & 1.3377 & 0.023 & 58.906 & 0.000 & [1.293, 1.382] \\
bs(total\_time\_per\_meal, degree=3, df=5)[2] & 1.4408 & 0.035 & 41.197 & 0.000 & [1.372, 1.509] \\
bs(total\_time\_per\_meal, degree=3, df=5)[3] & 1.8278 & 0.044 & 41.466 & 0.000 & [1.741, 1.914] \\
bs(total\_time\_per\_meal, degree=3, df=5)[4] & 2.6972 & 0.085 & 31.706 & 0.000 & [2.531, 2.864] \\
\bottomrule
\end{tabular}

\begin{tablenotes}\footnotesize
\item \textit{N} = 200,000, \quad
R\textsuperscript{2} = 0.297, \quad
Adj. R\textsuperscript{2} = 0.297, \quad
F-statistic: 1.408e+04 ($p < 0.001$), \quad
AIC: 5.907e+05, \quad
BIC: 5.908e+05.
\end{tablenotes}

\end{threeparttable}
\end{table}

\subsection{Implications for Prompt Engineers}

Spline-based insights help system designers identify where marginal returns flatten, guiding optimal resource allocation and reinforcing PCF’s adaptability mandate. Recognizing that excessive complexity may eventually undercut benefits prompts a key question: how do human or agentic prompt engineers find the ideal inflection point? Our data suggest iterative optimization, backed by adjustable SPARK settings, is the best route. By merging PCF’s scalability with non-linear modeling, prompt engineers can ensure agents remain coherent, cost-effective, and focused on user satisfaction.

\section{Discussion}

Our findings reveal that increasing configuration complexity enhances agent adaptability and performance, though these improvements plateau over time, as evidenced by meal-duration data across café tiers. This plateau effect has crucial implications for resource allocation - designers should focus optimization efforts on lower-complexity configurations where gains are substantial rather than over-engineering high-complexity systems. PCF addresses this challenge by leveraging combinatorial mathematics, category theory, topos theory, and rough fuzzy set theory to design agents tailored for complex, high-variability environments. However, the inherent constraints of large language models (LLMs), such as limited context windows and inference inefficiencies, highlight the critical need to fine-tune SPARK parameters.  Striking this balance is vital to maintaining resource efficiency while ensuring agents remain adaptable and robust\cite{hassan_hyper-dimensional_2022, fatima_rough_2024,kotyan_k_2024}.

At its heart, PCF represents a breakthrough in AI agent design, offering a structured framework for systematically aligning an agent’s attributes with the demands of dynamic environments. Whether applied to a bustling café, a high-pressure call center, a complex moon-based assembly line, or a home healthcare scenario, PCF enables agents to account for operational constraints and user-specific preferences with precision \cite{richard_jensen_computational_2008, gu_survey_2025, yuksel_multi-ai_2024,yin_goagent_2024,liu_-context_2023,wang_large_2024,street_llms_2024,zhou_self-discover_2024, song_trial_2024}. This adaptability is made possible through dynamic reconfiguration: agents leverage In-Context Learning (ICL) to adjust SPARK parameters in real time, avoiding the rigidity of fixed roles\cite{liu_-context_2023,wang_large_2024} . By integrating advanced mathematical tools such as functors and sheaves, PCF ensures logical consistency even as parameter complexity increases, allowing agents to remain responsive without sacrificing coherence.

Transparency is another cornerstone of PCF’s design. In high-stakes or regulated contexts, Explainable AI (XAI) plays a pivotal role in making agent behavior understandable and trustworthy. Emphasis on clarity is a core tenet of PCF, which produces agents that are explainable-by-design\cite{angelov_keynote_2021}. As researchers like Angelov argue \cite{angelov_keynote_2021}, true transparency should be an intrinsic property of a model’s structure, not solely a post-hoc analysis. Within PCF, an agent’s behavior is directly attributable to its human-readable SPARK configuration. This inherent transparency allows for straightforward debugging and auditing, fulfilling a core goal of Explainable AI (XAI)\cite{zhou_symbolic_2024,tull_towards_2024,jia_category-theoretical_2024,xiao_efficient_2024}.

PCF’s strengths extend seamlessly to multi-agent systems. By incorporating standardized communication protocols and shared knowledge representations, agents can share tasks, exchange domain-relevant knowledge, and adapt collectively to achieve complex goals. Tools from category theory, including morphisms, systematically define valid interactions, preventing role conflicts and ensuring smooth collaboration\cite{david_i_spivak_category_2014, tull_towards_2024, jia_category-theoretical_2024}. This approach is especially effective in scenarios requiring high levels of coordination, such as distributed sensor networks \cite{liu_graph_2024, amirkhani_consensus_2022} or co-creative design platforms\cite{lawton_drawing_2023}.

Emerging AI paradigms, such as Gödel agents \cite{yin_goagent_2024}, the Talker-Reasoner framework\cite{christakopoulou_agents_2024}, and Learn-by-interact\cite{su_learn-by-interact_2025}, further underscore the importance of abstract reasoning and transparent decision-making. PCF complements these innovations by introducing a combinatorial layer that tracks the evolution of SPARK parameters, enabling agents to maintain logical coherence while adapting dynamically to new inputs. By leveraging meta-learning across high-dimensional parameter spaces, PCF could ensure agents remain agile, avoiding computational inefficiencies and decision paralysis while meeting ethical and operational standards\cite{yin_goagent_2024, finn_model-agnostic_2017}.  

PCF introduces a foundational shift in how we design intelligent systems. Rather than relying on rigid, static configurations, it enables agents to dynamically reconfigure their behavioral parameters (Skills, Personalities, Approaches, Resources, and Knowledge) as situations evolve. This adaptability is not heuristic or ad hoc; it is grounded in formal mathematics. Category theory, topos logic, and sheaf structures ensure that even as agents transform, they maintain internal coherence and logical consistency.

Beyond refining individual agents, PCF unlocks a broader design paradigm for complex systems. Wherever outcomes depend on interacting variables, it provides a principled, tractable framework; transforming configuration from intuition-driven trial and error into rigorous exploration of possibility spaces. Its combinatorial architecture scales seamlessly from single-agent adaptation to the coordination of multi-agent ecosystems. Beyond single-agents and multi-agents, PCF can also optimize human-AI team composition by treating human skills as fixed parameters while dynamically configuring AI teammates to complement them in real time.

The same mathematical framework that ensures coherent agent behavior also supports systematic experimentation and design across domains. Developers can run ablation studies to quantify the impact of each SPARK dimension, revealing, for example, whether Personality or Approach most affects outcomes in a particular setting. Product teams can explore how different feature combinations influence user experience. Software architects can examine tradeoffs among modularity, performance, and maintainability within a structured parameter space.

PCF offers more than adaptability; it introduces a new standard of accountability in AI design. By making agent configurations explicit and adjustable, it exposes structural biases that traditional systems obscure. Static architectures embed unfairness beneath fixed rules. PCF reveals these patterns through formal transparency. Using structured simulations across diverse populations, developers can model millions of interactions, identify where disparities emerge, and tune SPARK parameters to better serve underrepresented groups. When unfairness arises, the path to correction is not speculative; it is computational and actionable.

This reframes fairness from a vague aspiration into a measurable design principle. The same mathematical structures that prevent contradictions within agents also illuminate how to build systems that adapt equitably across contexts. In PCF, fairness becomes an engineering property, not just a promise.

Crucially, PCF makes this adaptive intelligence deployable today. Using current LLMs, developers can define context, generate SPARK configurations, apply logical filters, and simulate behaviors, without retraining models or altering core architectures.

As AI systems evolve from tools into collaborators, the need for systems that can adapt with rigor and intention becomes paramount. PCF provides both the theoretical foundations and practical mechanisms to meet that need. The question is no longer "\textit{What can this AI do}?" but "\textit{What should this AI become, here and now}?". With PCF, that question has a structured, extensible, and deployable answer.

\section{Methods}

This section outlines the methodology used to simulate and analyze the behavior of PCF agents in complex environments. By employing parameterization and stochastic modeling, we recreated dynamic scenarios to evaluate how agents with varying SPARK configurations (Skills, Personalities, Approaches, Resources, and Knowledge) adapt and perform under diverse conditions.

\begin{figure}
    \centering
    \includegraphics[width=1\linewidth]{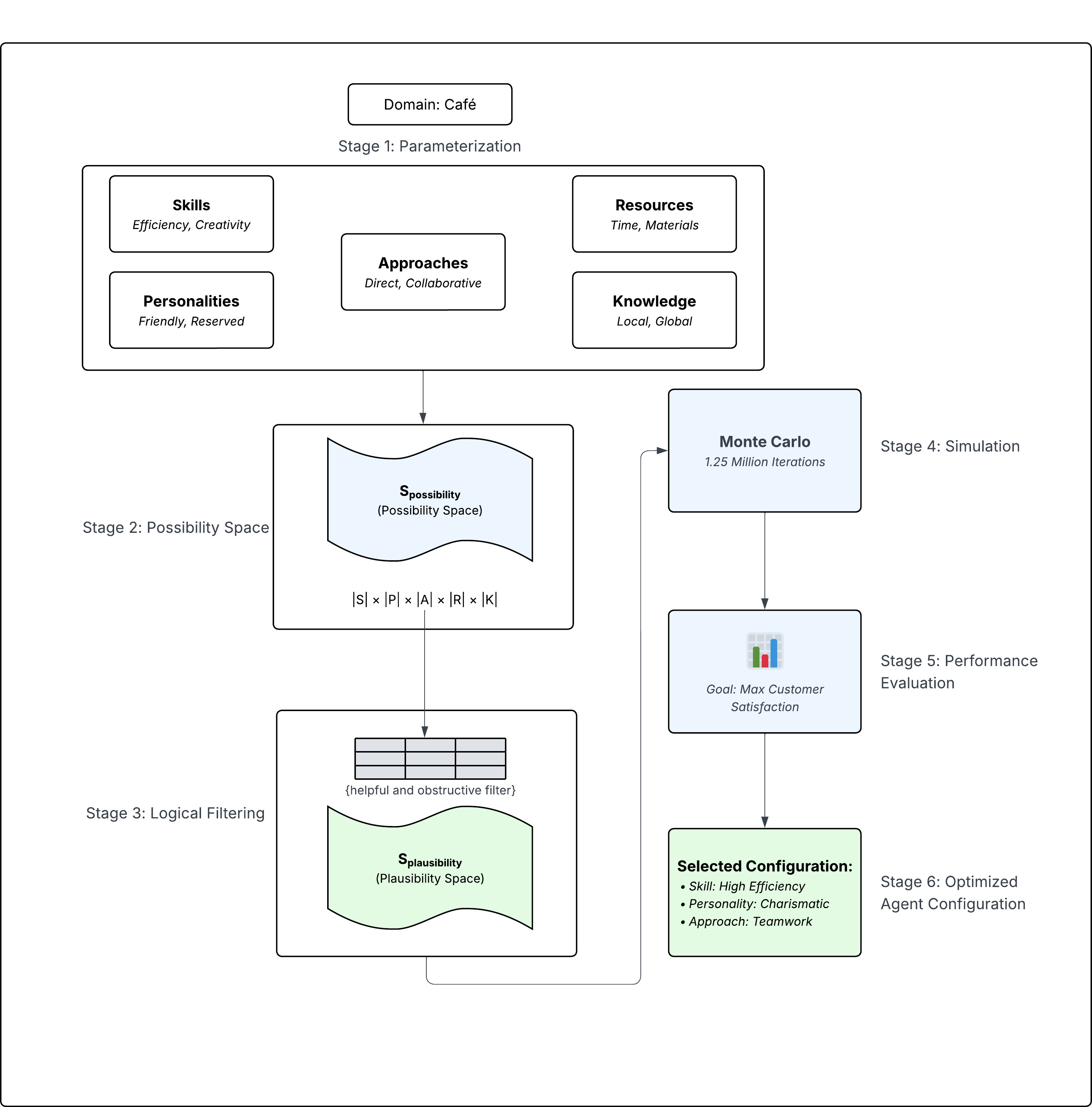}
    \caption{The Polymorphic Combinatorial Framework (PCF) Workflow. The process moves from LLM-assisted parameterization of the SPARK space to combinatorial expansion, followed by logical filtering to create a plausible set of agent configurations. These are then simulated and evaluated against predefined goals to identify the optimal agent configuration for a specific context. Note how logical filtering (Stage 3) eliminates contradictory combinations before simulation, ensuring all tested agents are coherent.}
    \label{fig4}
\end{figure}

\subsection{Simulation of Agent Behavior through Parameterization}

To evaluate the effectiveness of PCF, we developed comprehensive café simulations that model the full spectrum of service complexity found in real-world establishments. 

The café environment was deliberately chosen as our test domain because it provides a tractable yet sophisticated microcosm for complex real-world service ecosystems. This domain offers several critical advantages for validating PCF: (1) clear role hierarchies with interdependencies, (2) measurable performance outcomes, (3) varying complexity levels that scale naturally, and (4) direct transferability to high-stakes domains.
The structural parallels are compelling: a 'Chef' coordinating with a 'Sommelier' mirrors a 'Surgeon' collaborating with an 'Anesthesiologist'; 'customer satisfaction' directly parallels 'patient outcomes'; and the dynamic adaptation required for handling a 'rush order' shares fundamental properties with managing a 'technical emergency' in IT support. 

The five-tier complexity gradient (1-star to 5-star establishments) provides a controlled way to examine how agent configurations perform under increasing operational demands - from basic efficiency requirements in casual dining to sophisticated coordination needs in fine dining.
Critically, café operations involve the same core challenges that PCF addresses: agents must dynamically balance competing priorities (speed vs. quality), adapt their communication styles to different clientele, coordinate across specialized roles, and maintain performance under resource constraints. This makes café simulations an ideal proving ground for demonstrating PCF's ability to generate agents that remain coherent and effective as complexity scales.

Each café tier (1-5) was defined by specific parameter (SPARK) values assigned to roles such as Host, Server, and Cook, with additional specialized roles in higher-tier cafés, such as Sommelier and Mixologist. The full code for all scenarios is shared in the supplementary materials. Our methodology employed Claude for generating SPARK parameter configurations based on a gradient descent approach in prompt engineering. For each scenario, we manually and iteratively prompted to fine-tune environmental parameters such as resource availability, role-specific efficiencies, and client profiles. We filtered out outputs that exhibited logical inconsistencies or contradictions using structured consistency checks based on topos-theoretic principles. This approach allowed us to define key parameters for various roles and environmental factors, creating flexible and controlled simulations that reflected real-world complexities.

These systematically varied configurations allowed us to test the adaptability of PCF agents as they adjusted their SPARK attributes to meet evolving demands across different operational contexts.

 For each role, parameters, including new agents, were assigned values by Claude\cite{anthropic_introducing_nodate, qiao_autoact_2024, park_generative_2024, ahn_project_2024} on a scale from 1 to 10, representing dimensions such as skill level, training, and task complexity. For example, in one-star simulations, a parameter like $greeting\_efficiency$ for the Host role might be assigned a value of 2, reflecting minimal training and limited performance. In contrast, five-star simulations utilized higher parameter values to model highly skilled roles with intricate interdependencies, such as precise timing between Sommeliers and Cooks to deliver a seamless dining experience. These systematically varied configurations allowed us to test the adaptability of PCF agents as they adjusted their SPARK attributes to meet evolving demands.

As the complexity of the café tiers increased, so did the expectations (i.e. derived performance metrics and goals) for agent behavior. In one-star settings, speed, and efficiency were the primary goals, whereas in five-star environments, factors such as ambiance, personalized service, and multi-role coordination became critical. This gradient of complexity provided a structured yet flexible framework for analyzing how SPARK configurations influenced agent performance. By simulating a wide range of real-world scenarios, we were able to systematically assess the adaptability, scalability, and robustness of PCF agents under diverse conditions.

\subsection{Monte Carlo Simulation and Stochastic Modeling}
 
To evaluate AI performance across a wide range of scenarios while avoiding prohibitive computational costs, we developed the Monte Carlo Café Simulation. This simulation incorporated randomness to reflect the variability inherent in real-world AI agent outputs. Each simulation sampled a unique configuration of agent parameters, generating synthetic data that modeled customer interactions, agent performances, and environmental factors.

Monte Carlo methods enabled us to efficiently simulate 1.25 million iterations. This scale was necessary to capture the full variance across the five-dimensional SPARK space, with 250,000 iterations per café tier ensuring statistical robustness. This approach was both cost-effective and scalable, eliminating the need for real-world data collection while preserving the flexibility to model diverse conditions. The stochastic foundation of the simulation was informed by probabilistic processes inherent to LLMs, which exhibit variability during token generation and inference. Agent outputs were modeled as following a normal distribution, capturing the uncertainty and variability characteristic of LLM-driven behavior.

Agent performances and customer satisfaction scores were computed through weighted aggregations of key factors, including service quality, meal duration, cleanliness, and ambiance. The relative weight of each factor was dynamically adjusted across café tiers to reflect shifting priorities. For example, cleanliness played a more significant role in casual, one-star settings, while ambiance and personalized service were prioritized in five-star environments. These dynamic adjustments ensured the simulation accurately represented the evolving demands of different scenarios and reflected real world expectations and sociocultural norms.
By conducting numerous iterations, the Monte Carlo simulation produced a robust distribution of possible outcomes, enabling us to explore how agent configurations influenced performance under varying conditions. This systematic approach provided critical insights into the adaptability and effectiveness of PCF agents across a spectrum of real-world complexities.

\subsection{Data Analysis and Practical Implications}
The 1.25 million data points generated by the simulation were analyzed to investigate the relationships between agent complexity, performance metrics, and customer satisfaction. Clear trends emerged: as complexity increased across café tiers from one-star to five-star environments, both performance and satisfaction improved. However, this improvement followed a nonlinear trajectory, with diminishing marginal returns observed at higher complexity levels (refer to Figure 3). We find that prompt engineers exploring nuanced parameter-spaces through the PCF will produce more representative responses from the activation of parametric configurations optimized by use case.

For example, increased complexity yielded significant gains in satisfaction at lower tiers, while higher-tier environments benefit most from added sophistication. Additionally, at higher tiers, these gains plateaued, indicating a threshold beyond which additional complexity provides minimal benefit. This behavior mirrors phase transitions in physical systems, where systems shift between states but eventually reach equilibrium. Statistical analyses, including confidence intervals and standard errors, confirmed the reliability of these trends, with observed variances closely aligned with theoretical predictions (Table 3).

\begin{table}[ht!]
\centering
\caption{Statistics by Star Rating (Descriptive + 99\% Confidence Intervals)}
\label{tab:combined_stats_ci}
\small
\scriptsize
\begin{tabular}{l c c c c c c c c c c c c c c}
\toprule
\multirow{2}{*}{\textbf{Cafe}} 
 & \multicolumn{2}{c}{\textbf{Mean}}
 & \multicolumn{2}{c}{\textbf{Median}}
 & \multicolumn{2}{c}{\textbf{Min}}
 & \multicolumn{2}{c}{\textbf{Max}}
 & \multicolumn{2}{c}{\textbf{Std. Dev}}
 & \multicolumn{2}{c}{\textbf{99\% CI (Time)}}
 & \multicolumn{2}{c}{\textbf{99\% CI (Sat.)}} \\
\cmidrule(lr){2-3}
\cmidrule(lr){4-5}
\cmidrule(lr){6-7}
\cmidrule(lr){8-9}
\cmidrule(lr){10-11}
\cmidrule(lr){12-13}
\cmidrule(lr){14-15}
 & Time & Sat.
 & Time & Sat.
 & Time & Sat.
 & Time & Sat.
 & Time & Sat.
 & Lower & Upper
 & Lower & Upper \\
\midrule
5-Star 
 & 22.0249 & 6.5675  
 & 22      & 6.6     
 & 4       & 3.7333  
 & 40      & 9.3333  
 & 5.7431  & 0.6911  
 & 21.9953 & 22.0545 
 & 6.5639  & 6.571   
 \\
4-Star 
 & 16.4997 & 5.3738
 & 16      & 5.375
 & 3       & 1.3333
 & 30      & 9
 & 4.9739  & 0.9267
 & 16.4741 & 16.5253
 & 5.3691  & 5.3786
 \\
3-Star 
 & 16.4948 & 5.3759
 & 16      & 5.375
 & 3       & 1.4167
 & 30      & 9.625
 & 4.975   & 1.097
 & 16.4691 & 16.5204
 & 5.3702  & 5.3815
 \\
2-Star 
 & 10.9785 & 4.6852
 & 11      & 4.625
 & 2       & 0.75
 & 20      & 8.625
 & 4.0689  & 1.1898
 & 10.9575 & 10.9995
 & 4.679   & 4.6913
 \\
1-Star 
 & 10.9938 & 4.6868
 & 11      & 4.625
 & 2       & 0.75
 & 20      & 8.625
 & 4.0601  & 1.2945
 & 10.9729 & 11.0148
 & 4.6802  & 4.6935
 \\
\bottomrule
\end{tabular}
\end{table}

Visual representations of the data further illuminated these dynamics. Figures 1 and 2 demonstrated how shifts in agent efficiency and customer satisfaction correlated with rising complexity, while performance overlaps between adjacent tiers revealed that highly complex agents often achieved results comparable to simpler ones. This emphasizes the impact of stochastic variability, where probabilistic elements in agent behavior can blur distinctions between tiers.

\bibliographystyle{unsrtnat}
\bibliography{references}  

\begin{thebibliography}{43}
\providecommand{\natexlab}[1]{#1}
\providecommand{\url}[1]{\texttt{#1}}
\expandafter\ifx\csname urlstyle\endcsname\relax
  \providecommand{\doi}[1]{doi: #1}\else
  \providecommand{\doi}{doi: \begingroup \urlstyle{rm}\Url}\fi

\bibitem[Xi et~al.(2025)Xi, Chen, Guo, He, Ding, Hong, Zhang, Wang, Jin, Zhou, Zheng, Fan, Wang, Xiong, Zhou, Wang, Jiang, Zou, Liu, Yin, Dou, Weng, Cheng, Zhang, Qin, Zheng, Qiu, Huang, and Gui]{xi_rise_2025}
Zhiheng Xi, Wenxiang Chen, Xin Guo, Wei He, Yiwen Ding, Boyang Hong, Ming Zhang, Junzhe Wang, Senjie Jin, Enyu Zhou, Rui Zheng, Xiaoran Fan, Xiao Wang, Limao Xiong, Yuhao Zhou, Weiran Wang, Changhao Jiang, Yicheng Zou, Xiangyang Liu, Zhangyue Yin, Shihan Dou, Rongxiang Weng, Wensen Cheng, Qi~Zhang, Wenjuan Qin, Yongyan Zheng, Xipeng Qiu, Xuanjing Huang, and Tao Gui.
\newblock The {Rise} and {Potential} of {Large} {Language} {Model} {Based} {Agents}: {A} {Survey}.
\newblock \emph{Science China Information Sciences}, 68, January 2025.
\newblock \doi{https://doi.org/10.1007/s11432-024-4222-0}.
\newblock URL \url{https://link.springer.com/article/10.1007/s11432-024-4222-0}.

\bibitem[Pacheco and Carmo(2003)]{pacheco_role_2003}
Olga Pacheco and José Carmo.
\newblock A {Role} {Based} {Model} for the {Normative} {Specification} of {Organized} {Collective} {Agency} and {Agents} {Interaction}.
\newblock \emph{Autonomous Agents and Multi-Agent Systems}, 6\penalty0 (2):\penalty0 145--184, March 2003.
\newblock ISSN 1573-7454.
\newblock \doi{10.1023/A:1021884118023}.
\newblock URL \url{https://doi.org/10.1023/A:1021884118023}.

\bibitem[Zhou et~al.(2024{\natexlab{a}})Zhou, Ou, Ding, Li, Wu, Wang, Chen, Wang, Xu, Zhang, Chen, and Jiang]{zhou_symbolic_2024}
Wangchunshu Zhou, Yixin Ou, Shengwei Ding, Long Li, Jialong Wu, Tiannan Wang, Jiamin Chen, Shuai Wang, Xiaohua Xu, Ningyu Zhang, Huajun Chen, and Yuchen~Eleanor Jiang.
\newblock Symbolic {Learning} {Enables} {Self}-{Evolving} {Agents}, June 2024{\natexlab{a}}.
\newblock URL \url{http://arxiv.org/abs/2406.18532}.
\newblock arXiv:2406.18532 [cs] version: 1.

\bibitem[Pearl and Intriligator(2023)]{pearl_persona_2023}
David Pearl and James Intriligator.
\newblock Persona {Multiplication}: {A} {Method} to {Avoid} {Designed} {Injustice}.
\newblock \emph{The International Journal of Design in Society}, 17\penalty0 (1):\penalty0 31--44, 2023.
\newblock ISSN 2325-1328, 2325-1360.
\newblock \doi{10.18848/2325-1328/CGP/v17i01/31-44}.
\newblock URL \url{https://cgscholar.com/bookstore/works/persona-multiplication}.

\bibitem[Kong et~al.(2007)Kong, McCullagh, Meng, and Nicolae]{nair_further_2007}
Augustine Kong, Peter McCullagh, Xiao-Li Meng, and Dan~L. Nicolae.
\newblock {FURTHER} {EXPLORATIONS} {OF} {LIKELIHOOD} {THEORY} {FOR} {MONTE} {CARLO} {INTEGRATION}.
\newblock In \emph{Advances in {Statistical} {Modeling} and {Inference}}, pages 563--592. WORLD SCIENTIFIC, March 2007.
\newblock ISBN 978-981-270-369-9 978-981-270-829-8.
\newblock \doi{10.1142/9789812708298_0028}.
\newblock URL \url{http://www.worldscientific.com/doi/abs/10.1142/9789812708298_0028}.

\bibitem[Bender et~al.(2021)Bender, Gebru, McMillan-Major, and Shmitchell]{bender_dangers_2021}
Emily~M. Bender, Timnit Gebru, Angelina McMillan-Major, and Shmargaret Shmitchell.
\newblock On the {Dangers} of {Stochastic} {Parrots}: {Can} {Language} {Models} {Be} {Too} {Big}?
\newblock In \emph{Proceedings of the 2021 {ACM} {Conference} on {Fairness}, {Accountability}, and {Transparency}}, {FAccT} '21, pages 610--623, New York, NY, USA, March 2021. Association for Computing Machinery.
\newblock ISBN 978-1-4503-8309-7.
\newblock \doi{10.1145/3442188.3445922}.
\newblock URL \url{https://dl.acm.org/doi/10.1145/3442188.3445922}.

\bibitem[Selig et~al.(2012)Selig, Oppermann, and Enßlin]{selig_improving_2012}
Marco Selig, Niels Oppermann, and Torsten~A. Enßlin.
\newblock Improving stochastic estimates with inference methods: {Calculating} matrix diagonals.
\newblock \emph{Physical Review E}, 85\penalty0 (2):\penalty0 021134, February 2012.
\newblock ISSN 1539-3755, 1550-2376.
\newblock \doi{10.1103/PhysRevE.85.021134}.
\newblock URL \url{https://link.aps.org/doi/10.1103/PhysRevE.85.021134}.

\bibitem[Ramalho et~al.(2013)Ramalho, Selig, Gerland, and Enßlin]{ramalho_simulation_2013}
Tiago Ramalho, Marco Selig, Ulrich Gerland, and Torsten~A. Enßlin.
\newblock Simulation of stochastic network dynamics via entropic matching.
\newblock \emph{Physical Review E}, 87\penalty0 (2):\penalty0 022719, February 2013.
\newblock ISSN 1539-3755, 1550-2376.
\newblock \doi{10.1103/PhysRevE.87.022719}.
\newblock URL \url{https://link.aps.org/doi/10.1103/PhysRevE.87.022719}.

\bibitem[Bar~Massada and Carmel(2008)]{bar_massada_incorporating_2008}
Avi Bar~Massada and Yohay Carmel.
\newblock Incorporating output variance in local sensitivity analysis for stochastic models.
\newblock \emph{Ecological Modelling}, 213\penalty0 (3):\penalty0 463--467, May 2008.
\newblock ISSN 0304-3800.
\newblock \doi{10.1016/j.ecolmodel.2008.01.021}.
\newblock URL \url{https://www.sciencedirect.com/science/article/pii/S0304380008000550}.

\bibitem[Pryzant et~al.(2020)Pryzant, Diehl~Martinez, Dass, Kurohashi, Jurafsky, and Yang]{pryzant_automatically_2020}
Reid Pryzant, Richard Diehl~Martinez, Nathan Dass, Sadao Kurohashi, Dan Jurafsky, and Diyi Yang.
\newblock Automatically {Neutralizing} {Subjective} {Bias} in {Text}.
\newblock \emph{Proceedings of the AAAI Conference on Artificial Intelligence}, 34\penalty0 (01):\penalty0 480--489, April 2020.
\newblock ISSN 2374-3468, 2159-5399.
\newblock \doi{10.1609/aaai.v34i01.5385}.
\newblock URL \url{https://ojs.aaai.org/index.php/AAAI/article/view/5385}.

\bibitem[Baez(2021)]{baez_topos_2021}
John Baez.
\newblock Topos {Theory} in a {Nutshell}, October 2021.
\newblock URL \url{https://math.ucr.edu/home/baez/topos.html}.

\bibitem[{David I. Spivak}(2014)]{david_i_spivak_category_2014}
{David I. Spivak}.
\newblock \emph{Category {Theory} for the {Sciences}}.
\newblock 2014.
\newblock URL \url{http://archive.org/details/cattheory}.

\bibitem[Tull et~al.(2024)Tull, Lorenz, Clark, Khan, and Coecke]{tull_towards_2024}
Sean Tull, Robin Lorenz, Stephen Clark, Ilyas Khan, and Bob Coecke.
\newblock Towards {Compositional} {Interpretability} for {XAI}, June 2024.
\newblock URL \url{http://arxiv.org/abs/2406.17583}.
\newblock arXiv:2406.17583 [cs].

\bibitem[{Anthropic}()]{anthropic_introducing_nodate}
{Anthropic}.
\newblock Introducing computer use, a new {Claude} 3.5 {Sonnet}, and {Claude} 3.5 {Haiku}.
\newblock URL \url{https://www.anthropic.com/news/3-5-models-and-computer-use}.

\bibitem[Hassan et~al.(2022)Hassan, Halawani, Mohammad, and Saleh]{hassan_hyper-dimensional_2022}
Eman Hassan, Yasmin Halawani, Baker Mohammad, and Hani Saleh.
\newblock Hyper-{Dimensional} {Computing} {Challenges} and {Opportunities} for {AI} {Applications}.
\newblock \emph{IEEE Access}, 10:\penalty0 97651--97664, 2022.
\newblock ISSN 2169-3536.
\newblock \doi{10.1109/ACCESS.2021.3059762}.
\newblock URL \url{https://ieeexplore.ieee.org/abstract/document/9354795}.
\newblock Conference Name: IEEE Access.

\bibitem[Woolsey et~al.(2020)Woolsey, Chen, and Ji]{woolsey_combinatorial_2020}
Nicholas Woolsey, Rong-Rong Chen, and Mingyue Ji.
\newblock A {Combinatorial} {Design} for {Cascaded} {Coded} {Distributed} {Computing} on {General} {Networks}, August 2020.
\newblock URL \url{http://arxiv.org/abs/2008.00581}.
\newblock arXiv:2008.00581 [cs, math].

\bibitem[{Richard Jensen} and {Qiang Shen}(2008)]{richard_jensen_computational_2008}
{Richard Jensen} and {Qiang Shen}.
\newblock Computational {Intelligence} and {Feature} {Selection}: {Rough} and {Fuzzy} {Approaches} {\textbar} {IEEE} {eBooks} {\textbar} {IEEE} {Xplore}, 2008.
\newblock URL \url{https://ieeexplore-ieee-org.ezproxy.library.tufts.edu/book/5236578}.

\bibitem[Fatima and Javaid(2024)]{fatima_rough_2024}
Abeer Fatima and Imran Javaid.
\newblock Rough set theory applied to finite dimensional vector spaces.
\newblock \emph{Information Sciences}, 659:\penalty0 120072, February 2024.
\newblock ISSN 0020-0255.
\newblock \doi{10.1016/j.ins.2023.120072}.
\newblock URL \url{https://www.sciencedirect.com/science/article/pii/S0020025523016584}.

\bibitem[Halmos(1974)]{halmos_naive_1974}
Paul~Richard Halmos.
\newblock \emph{Naive set theory}.
\newblock Undergraduate texts in mathematics. Springer, New York Berlin Paris [etc.], 1974.
\newblock ISBN 978-0-387-90092-6 978-3-540-90092-4.

\bibitem[{Julien Chaumond} et~al.(2024){Julien Chaumond}, {Hamza Tahir}, and {Antonio Gulli}]{julien_chaumond_llm_2024}
{Julien Chaumond}, {Hamza Tahir}, and {Antonio Gulli}.
\newblock \emph{{LLM} {Engineer}'s {Handbook}}.
\newblock 2024.

\bibitem[Mingard et~al.(2025)Mingard, Rees, Valle-Pérez, and Louis]{mingard_deep_2025}
Chris Mingard, Henry Rees, Guillermo Valle-Pérez, and Ard~A. Louis.
\newblock Deep neural networks have an inbuilt {Occam}’s razor.
\newblock \emph{Nature Communications}, 16\penalty0 (1):\penalty0 220, January 2025.
\newblock ISSN 2041-1723.
\newblock \doi{10.1038/s41467-024-54813-x}.
\newblock URL \url{https://www.nature.com/articles/s41467-024-54813-x}.
\newblock Publisher: Nature Publishing Group.

\bibitem[Jia et~al.(2024)Jia, Peng, Yang, and Chen]{jia_category-theoretical_2024}
Yiyang Jia, Guohong Peng, Zheng Yang, and Tianhao Chen.
\newblock Category-{Theoretical} and {Topos}-{Theoretical} {Frameworks} in {Machine} {Learning}: {A} {Survey}, August 2024.
\newblock URL \url{http://arxiv.org/abs/2408.14014}.
\newblock arXiv:2408.14014 [cs].

\bibitem[Nezhurina et~al.(2024)Nezhurina, Cipolina-Kun, Cherti, and Jitsev]{nezhurina_alice_2024}
Marianna Nezhurina, Lucia Cipolina-Kun, Mehdi Cherti, and Jenia Jitsev.
\newblock Alice in {Wonderland}: {Simple} {Tasks} {Showing} {Complete} {Reasoning} {Breakdown} in {State}-{Of}-the-{Art} {Large} {Language} {Models}, June 2024.
\newblock URL \url{http://arxiv.org/abs/2406.02061}.
\newblock arXiv:2406.02061 [cs].

\bibitem[Gu et~al.(2025)Gu, Jiang, Shi, Tan, Zhai, Xu, Li, Shen, Ma, Liu, Wang, and Guo]{gu_survey_2025}
Jiawei Gu, Xuhui Jiang, Zhichao Shi, Hexiang Tan, Xuehao Zhai, Chengjin Xu, Wei Li, Yinghan Shen, Shengjie Ma, Honghao Liu, Yuanzhuo Wang, and Jian Guo.
\newblock A {Survey} on {LLM}-as-a-{Judge}, January 2025.
\newblock URL \url{http://arxiv.org/abs/2411.15594}.
\newblock arXiv:2411.15594 [cs].

\bibitem[Kotyan et~al.(2024)Kotyan, Ueda, and Vargas]{kotyan_k_2024}
Shashank Kotyan, Tatsuya Ueda, and Danilo~Vasconcellos Vargas.
\newblock k* {Distribution}: {Evaluating} the {Latent} {Space} of {Deep} {Neural} {Networks} using {Local} {Neighborhood} {Analysis}, August 2024.
\newblock URL \url{http://arxiv.org/abs/2312.04024}.
\newblock arXiv:2312.04024 [cs].

\bibitem[Yuksel and Sawaf(2024)]{yuksel_multi-ai_2024}
Kamer~Ali Yuksel and Hassan Sawaf.
\newblock A {Multi}-{AI} {Agent} {System} for {Autonomous} {Optimization} of {Agentic} {AI} {Solutions} via {Iterative} {Refinement} and {LLM}-{Driven} {Feedback} {Loops}, December 2024.
\newblock URL \url{http://arxiv.org/abs/2412.17149}.
\newblock arXiv:2412.17149 [cs].

\bibitem[Yin et~al.(2024)Yin, Wang, Pan, Wan, and Wang]{yin_goagent_2024}
Xunjian Yin, Xinyi Wang, Liangming Pan, Xiaojun Wan, and William~Yang Wang.
\newblock Gödel {Agent}: {A} {Self}-{Referential} {Agent} {Framework} for {Recursive} {Self}-{Improvement}, October 2024.
\newblock URL \url{http://arxiv.org/abs/2410.04444}.
\newblock arXiv:2410.04444 version: 2.

\bibitem[Liu et~al.(2023)Liu, Ye, Xing, and Zou]{liu_-context_2023}
Sheng Liu, Haotian Ye, Lei Xing, and James Zou.
\newblock In-context {Vectors}: {Making} {In} {Context} {Learning} {More} {Effective} and {Controllable} {Through} {Latent} {Space} {Steering}, November 2023.
\newblock URL \url{https://arxiv.org/abs/2311.06668v3}.

\bibitem[Wang et~al.(2024)Wang, Chen, Wang, U, Li, and Guo]{wang_large_2024}
Xin Wang, Zirui Chen, Haofen Wang, Leong~Hou U, Zhao Li, and Wenbin Guo.
\newblock Large {Language} {Model} {Enhanced} {Knowledge} {Representation} {Learning}: {A} {Survey}, July 2024.
\newblock URL \url{http://arxiv.org/abs/2407.00936}.
\newblock arXiv:2407.00936 [cs].

\bibitem[Street et~al.(2024)Street, Siy, Keeling, Baranes, Barnett, McKibben, Kanyere, Lentz, Arcas, and Dunbar]{street_llms_2024}
Winnie Street, John~Oliver Siy, Geoff Keeling, Adrien Baranes, Benjamin Barnett, Michael McKibben, Tatenda Kanyere, Alison Lentz, Blaise Aguera~y Arcas, and Robin I.~M. Dunbar.
\newblock {LLMs} achieve adult human performance on higher-order theory of mind tasks, May 2024.
\newblock URL \url{http://arxiv.org/abs/2405.18870}.
\newblock arXiv:2405.18870 [cs].

\bibitem[Zhou et~al.(2024{\natexlab{b}})Zhou, Pujara, Ren, Chen, Cheng, Le, Chi, Zhou, Mishra, and Zheng]{zhou_self-discover_2024}
Pei Zhou, Jay Pujara, Xiang Ren, Xinyun Chen, Heng-Tze Cheng, Quoc~V. Le, Ed~H. Chi, Denny Zhou, Swaroop Mishra, and Huaixiu~Steven Zheng.
\newblock Self-{Discover}: {Large} {Language} {Models} {Self}-{Compose} {Reasoning} {Structures}, February 2024{\natexlab{b}}.
\newblock URL \url{http://arxiv.org/abs/2402.03620}.
\newblock arXiv:2402.03620 [cs].

\bibitem[Song et~al.(2024)Song, Yin, Yue, Huang, Li, and Lin]{song_trial_2024}
Yifan Song, Da~Yin, Xiang Yue, Jie Huang, Sujian Li, and Bill~Yuchen Lin.
\newblock Trial and {Error}: {Exploration}-{Based} {Trajectory} {Optimization} for {LLM} {Agents}, 2024.
\newblock URL \url{https://arxiv.org/abs/2403.02502}.
\newblock Version Number: 2.

\bibitem[Angelov(2021)]{angelov_keynote_2021}
Plamen Angelov.
\newblock Keynote: {Explainable}-by-design {Deep} {Learning}.
\newblock In \emph{2021 {IEEE} {International} {Conference} on {Pervasive} {Computing} and {Communications} {Workshops} and other {Affiliated} {Events} ({PerCom} {Workshops})}, pages 175--175, March 2021.
\newblock \doi{10.1109/PerComWorkshops51409.2021.9431114}.
\newblock URL \url{https://ieeexplore.ieee.org/document/9431114}.

\bibitem[Xiao et~al.(2024)Xiao, Tian, Chen, Han, and Lewis]{xiao_efficient_2024}
Guangxuan Xiao, Yuandong Tian, Beidi Chen, Song Han, and Mike Lewis.
\newblock Efficient {Streaming} {Language} {Models} with {Attention} {Sinks}, April 2024.
\newblock URL \url{http://arxiv.org/abs/2309.17453}.
\newblock arXiv:2309.17453 [cs].

\bibitem[Liu et~al.(2024)Liu, Zhang, Shi, Liu, Niyato, Ai, and Shen]{liu_graph_2024}
Ziheng Liu, Jiayi Zhang, Enyu Shi, Zhilong Liu, Dusit Niyato, Bo~Ai, and Xuemin Shen.
\newblock Graph {Neural} {Network} {Meets} {Multi}-{Agent} {Reinforcement} {Learning}: {Fundamentals}, {Applications}, and {Future} {Directions}.
\newblock \emph{IEEE Wireless Communications}, 31\penalty0 (6):\penalty0 39--47, December 2024.
\newblock ISSN 1558-0687.
\newblock \doi{10.1109/MWC.015.2300595}.
\newblock URL \url{https://ieeexplore.ieee.org/abstract/document/10638531}.
\newblock Conference Name: IEEE Wireless Communications.

\bibitem[Amirkhani and Barshooi(2022)]{amirkhani_consensus_2022}
Abdollah Amirkhani and Amir~Hossein Barshooi.
\newblock Consensus in multi-agent systems: a review.
\newblock \emph{Artificial Intelligence Review}, 55\penalty0 (5):\penalty0 3897--3935, June 2022.
\newblock ISSN 1573-7462.
\newblock \doi{10.1007/s10462-021-10097-x}.
\newblock URL \url{https://doi.org/10.1007/s10462-021-10097-x}.

\bibitem[Lawton et~al.(2023)Lawton, Ibarrola, Ventura, and Grace]{lawton_drawing_2023}
Tomas Lawton, Francisco~J Ibarrola, Dan Ventura, and Kazjon Grace.
\newblock Drawing with {Reframer}: {Emergence} and {Control} in {Co}-{Creative} {AI}.
\newblock In \emph{Proceedings of the 28th {International} {Conference} on {Intelligent} {User} {Interfaces}}, {IUI} '23, pages 264--277, New York, NY, USA, March 2023. Association for Computing Machinery.
\newblock ISBN 9798400701061.
\newblock \doi{10.1145/3581641.3584095}.
\newblock URL \url{https://dl.acm.org/doi/10.1145/3581641.3584095}.

\bibitem[Christakopoulou et~al.(2024)Christakopoulou, Mourad, and Matarić]{christakopoulou_agents_2024}
Konstantina Christakopoulou, Shibl Mourad, and Maja Matarić.
\newblock Agents {Thinking} {Fast} and {Slow}: {A} {Talker}-{Reasoner} {Architecture}, October 2024.
\newblock URL \url{http://arxiv.org/abs/2410.08328}.
\newblock arXiv:2410.08328.

\bibitem[Su et~al.(2025)Su, Sun, Yoon, Yin, Yu, and Arık]{su_learn-by-interact_2025}
Hongjin Su, Ruoxi Sun, Jinsung Yoon, Pengcheng Yin, Tao Yu, and Sercan~Ö Arık.
\newblock Learn-by-interact: {A} {Data}-{Centric} {Framework} for {Self}-{Adaptive} {Agents} in {Realistic} {Environments}, January 2025.
\newblock URL \url{http://arxiv.org/abs/2501.10893}.
\newblock arXiv:2501.10893 [cs].

\bibitem[Finn et~al.(2017)Finn, Abbeel, and Levine]{finn_model-agnostic_2017}
Chelsea Finn, Pieter Abbeel, and Sergey Levine.
\newblock Model-{Agnostic} {Meta}-{Learning} for {Fast} {Adaptation} of {Deep} {Networks}.
\newblock Proceedings of {Machine} {Learning} {Research}. PMLR, July 2017.
\newblock URL \url{https://proceedings.mlr.press/v70/finn17a.html}.

\bibitem[Qiao et~al.(2024)Qiao, Zhang, Fang, Luo, Zhou, Jiang, Lv, and Chen]{qiao_autoact_2024}
Shuofei Qiao, Ningyu Zhang, Runnan Fang, Yujie Luo, Wangchunshu Zhou, Yuchen~Eleanor Jiang, Chengfei Lv, and Huajun Chen.
\newblock {AutoAct}: {Automatic} {Agent} {Learning} from {Scratch} for {QA} via {Self}-{Planning}, 2024.
\newblock URL \url{https://arxiv.org/abs/2401.05268}.
\newblock Version Number: 4.

\bibitem[Park et~al.(2024)Park, Zou, Shaw, Hill, Cai, Morris, Willer, Liang, and Bernstein]{park_generative_2024}
Joon~Sung Park, Carolyn~Q. Zou, Aaron Shaw, Benjamin~Mako Hill, Carrie Cai, Meredith~Ringel Morris, Robb Willer, Percy Liang, and Michael~S. Bernstein.
\newblock Generative {Agent} {Simulations} of 1,000 {People}, November 2024.
\newblock URL \url{http://arxiv.org/abs/2411.10109}.
\newblock arXiv:2411.10109.

\bibitem[Ahn et~al.(2024)Ahn, Becker, Carroll, Christie, Cortes, Demirci, Du, Li, Luo, Wang, Willows, Yang, and Yang]{ahn_project_2024}
Andrew Ahn, Nic Becker, Stephanie Carroll, Nico Christie, Manuel Cortes, Arda Demirci, Melissa Du, Frankie Li, Shuying Luo, Peter~Y. Wang, Mathew Willows, Feitong Yang, and Guangyu~Robert Yang.
\newblock Project {Sid}: {Many}-agent simulations toward {AI} civilization, October 2024.
\newblock URL \url{http://arxiv.org/abs/2411.00114}.
\newblock arXiv:2411.00114 [cs].

\end{thebibliography}






\end{document}